\documentclass[10pt,twocolumn,letterpaper]{article}

\usepackage[pagenumbers]{cvpr} 

\usepackage[dvipsnames]{xcolor}

\definecolor{cvprblue}{rgb}{0.21,0.49,0.74}
\usepackage[pagebackref,breaklinks,colorlinks,citecolor=cvprblue]{hyperref}

\usepackage{float}
\usepackage{stfloats}
\usepackage{enumerate}

\usepackage{svg}
\usepackage{amsmath}

\graphicspath{{Images/}}

\def\paperID{3} 
\def\confName{CVPR}
\def\confYear{2024}

\title{EucliDreamer: Fast and High-Quality Texturing for 3D Models with Depth-Conditioned Stable Diffusion}

\author{Cindy Le\\
Columbia University\\
{\tt\small xl2738@columbia.edu}
\and
Congrui Hetang\\
Carnegie Mellon University\\
{\tt\small congruihetang@gmail.com}
\and
Chendi Lin\\
Carnegie Mellon University\\
{\tt\small chendil@alumni.cmu.edu}
\and
Ang Cao\\
University of Michigan\\
{\tt\small ancao@umich.edu}
\and
Yihui He\\
Carnegie Mellon University\\
{\tt\small he2@alumni.cmu.edu}
}

\usepackage{listings}
\usepackage{xcolor}

\begin{document}
\twocolumn[{%
\maketitle
\begin{center}
    \centering
    \includegraphics[width=0.8\linewidth]{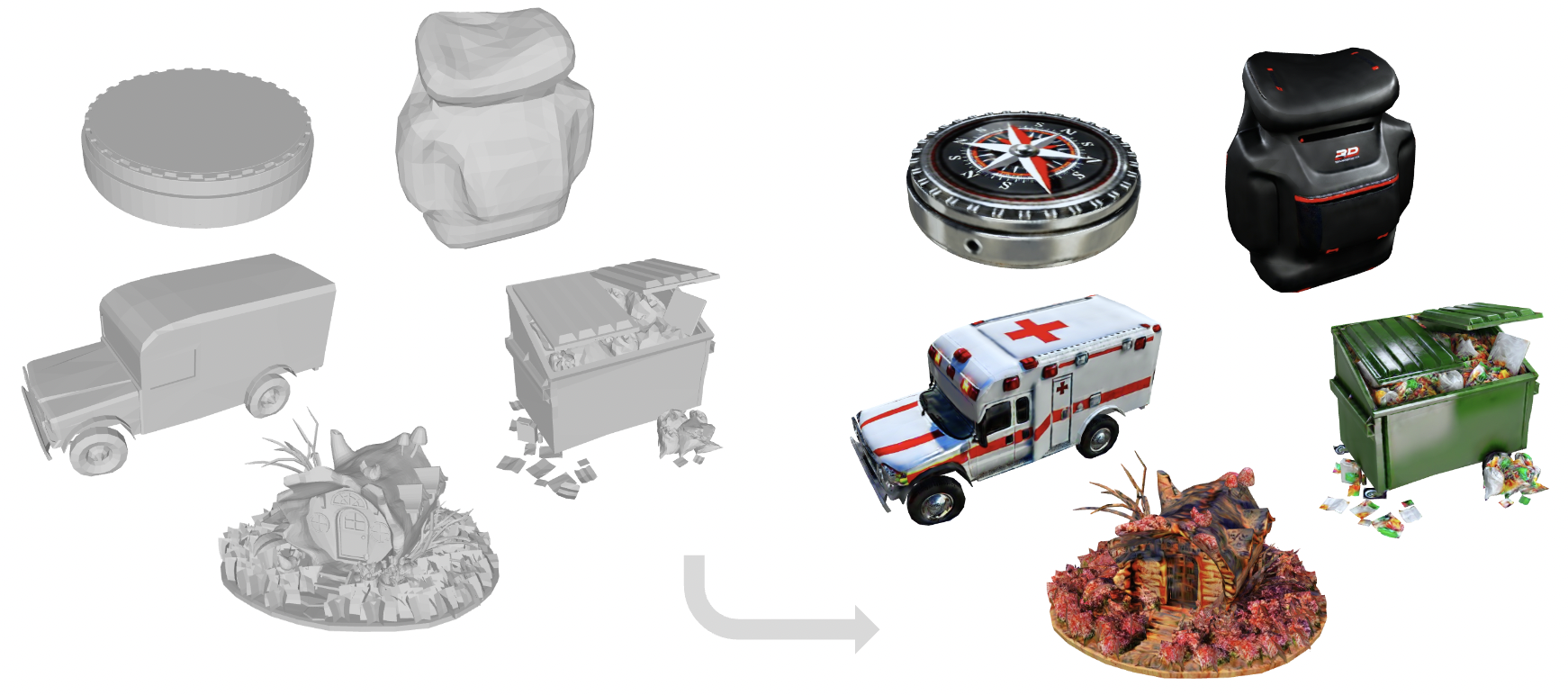}
    \captionof{figure}{3D objects textured by EucliDreamer. The generated textures are realistic and highly detailed.}
    \label{fig:cover}
\end{center}
}]

\begin{abstract}

We present EucliDreamer \cite{congruiEucli2023}, a simple and effective method to generate textures for 3D models given text prompts and meshes. The texture is parametrized as an implicit function on the 3D surface, which is optimized with the Score Distillation Sampling (SDS) process~\cite{poole2022dreamfusion} and differentiable rendering~\cite{nvdiffrast2020}. To generate high-quality textures, we leverage a depth-conditioned Stable Diffusion~\cite{Rombach_2022_CVPR} model guided by the depth image rendered from the mesh. We test our approach on 3D models in Objaverse~\cite{deitke2022objaverse} and conducted a user study, which shows its superior quality compared to existing texturing methods like Text2Tex. In addition, our method converges 2 times faster than DreamFusion. Through text prompting, textures of diverse art styles can be produced. We hope Euclidreamer proides a viable solution to automate a labor-intensive stage in 3D content creation.


\end{abstract}  




\section{Method}
\label{sec:method}

\begin{figure*} [h!]
    \centering
    \includegraphics[width=0.8\linewidth]{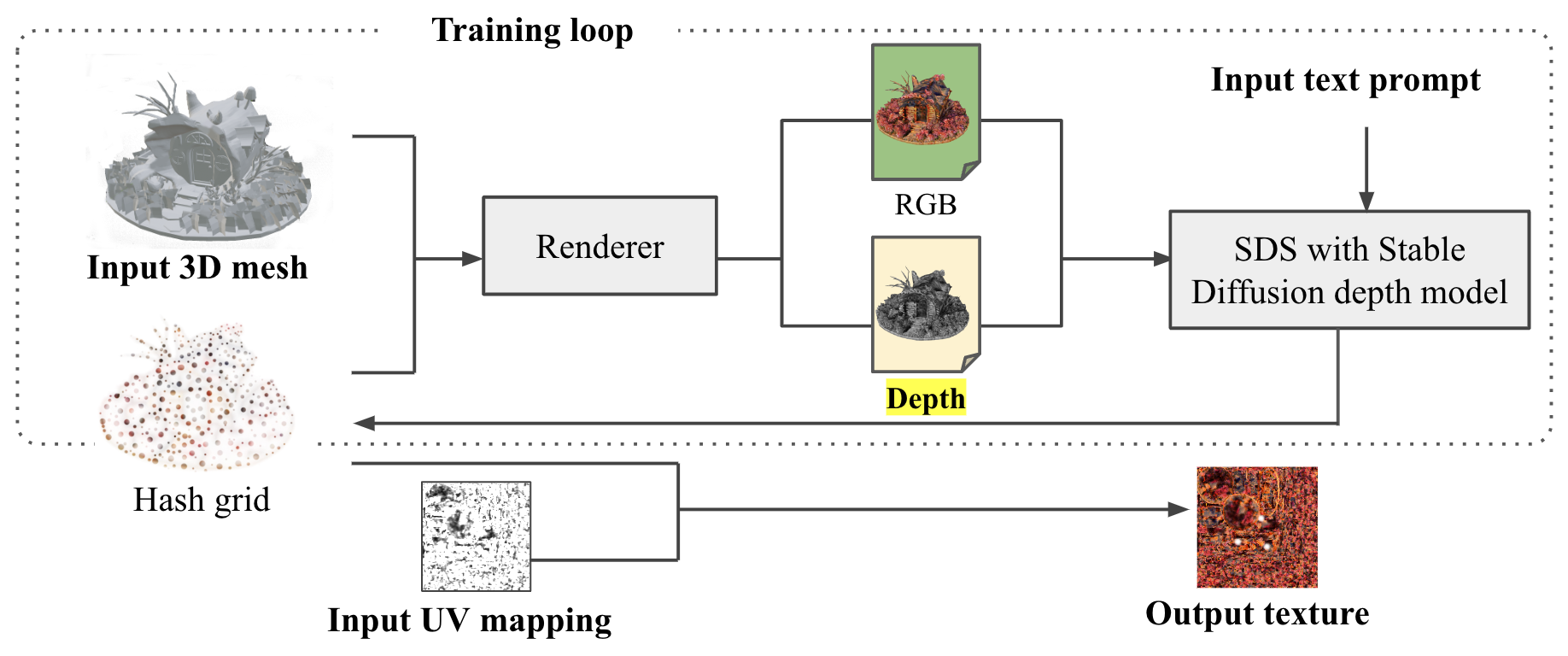}
    \caption{\textbf{Our method.} Multiple views of the texured mesh are produced by a differentiable renderer, then the SDS loss will back-propagate to update a hash grid, conditioned on text prompts and depth images.}
    \label{fig:method}
\end{figure*}

Our method is illustrated in Figure ~\ref{fig:method}. Given a 3D mesh, we represent its texture as an implicit function defined on its surface - an MLP mapping from 2D coordinates on the surface to RGB color values, resembling a NeRF ~\cite{mildenhall2020nerf}. For compact representation and faster inference, hash grid ~\cite{mueller2022instant} is adopted in our implementation. 

During training, the parameterized texture is optimized to encode a high-quality texture corresponding to a text prompt. Inspired by DreamFusion ~\cite{poole2022dreamfusion}, we acquire supervision signal from a high-capacity text-to-image diffusion model, such as Stable Diffusion. Specifically, the texture is randomly initialized in the beginning. In each training step, 2D images of the textured mesh are produced by a differentiable renderer from random view points, and are iteratively refined via the score distillation sampling (SDS) process by the text-to-image diffusion model. The SDS loss backprops to the MLP through the differentiable renderer.

To improve texture quality and speed-up convergence, we introduce additional depth conditioning for the text-to-image diffusion model, as depth information removes a lot of ambiguities, and helps enhance cross-view consistency. To this end, we leverage a version of Stable Diffusion model that conditions on a depth image ~\footnote{\url{https://huggingface.co/stabilityai/stable-diffusion-2-depth}} in addition to text. During training, the depth image is acquired by rendering from the 3D mesh. 

When the optimization completes, the texture can be exported by querying each point on the 3D surface. It is then stored as a 2D image and a UV map, suitable for further rendering or editing with ease.

\section{Experiment}
\label{sec:ablation}

We use 3D models from Objaverse~\cite{deitke2022objaverse} (a large open-sourced dataset of objects with 800K+ textured models with descriptive captions and tags) to test our approach.

For the main experiments, we run EucliDreamer on a set of 3D meshes from Objaverse. Qualitatively, we compare EucliDreamer visually with other 3D texturing methods, including Text2Tex~\cite{chen2023text2tex}, CLIPMesh~\cite{Mohammad_Khalid_2022}, and Latent-Paint~\cite{metzer2022latentnerf}. Quantitatively, we conduct a user study involving 28 professionals in 3D content creation. They are asked to rank the textures generated by different methods in terms of quality.

Additionally, we demonstrate that EucliDreamer can easily generate diverse art styles for the same object via text prompting. Due to page limit, these results are shown in Figure ~\ref{fig:styles} in supplementary materials.

The hyperparameters of the system is selected to optimize the generation quality. Due to page limit, the detailed study of the effect of these parameters and other implementation details can be found in Supplementary Materials ~\ref{sec:suppl}. For all experiments, we use concise text prompts to specify the content and style, e.g. "3D rendering of a [description of the object], realistic, high-quality".







\begin{figure*} [h!]
    \centering
    \includegraphics[width=0.75\linewidth]{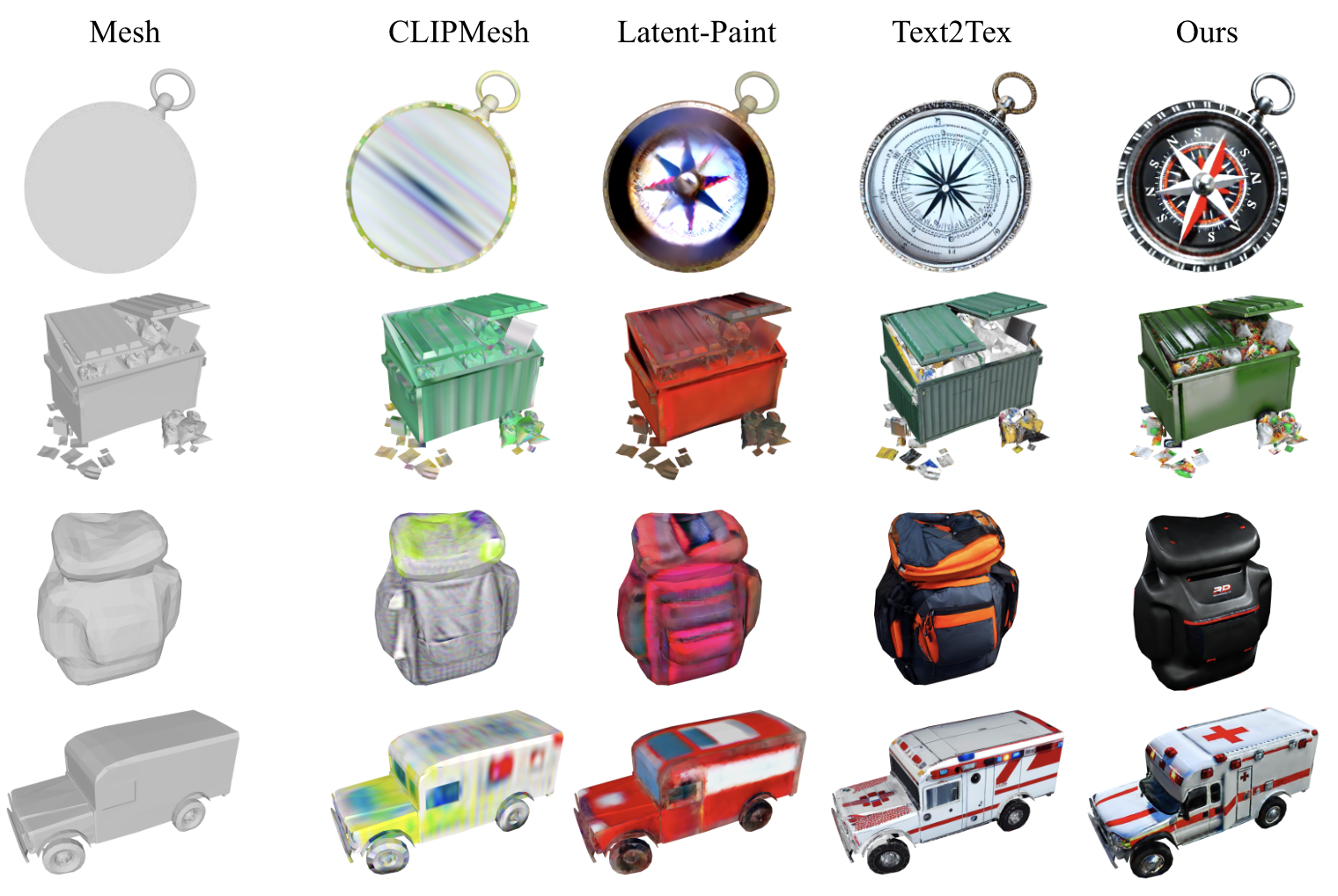}
    \caption{\textbf{Comparisons with existing texturing methods.} Four objects are used for illustration from Objaverse~\cite{deitke2022objaverse}. The rendering performance of the first three methods, CLIPMesh, Latent-Paint, and Text2Tex are discussed in \cite{chen2023text2tex}. Overall, the examples demonstrate a clear win of EucliDreamer over the baselines ~\cite{Mohammad_Khalid_2022, metzer2022latentnerf, chen2023text2tex} in terms of clarity, sharpness and level of details.}
    \label{fig:result}
\end{figure*}



\section{Results}
\label{sec:results}

\subsection{Visual Benchmark on Objaverse Models}

Figure \ref{fig:result} shows some qualitative results. Overall, the generated textures are of superior quality in terms of the realism, the level of detail, cross-view consistency and aesthetics. Notice on the dumpster model, EucliDreamer even accurately captured the specular pattern of the metallic material, a special trait resembling texture baking. For the compass model, not only the coloring is realistic, it also accurately recovers the details such as the characters and the small decorations on the rim.



\subsection{User Study on Generation Quality}

We invited 28 participants for the user study who are professionals in game industry and 3D content creation, including engineers, researchers and artist. Most of them are experienced with the 3D modeling workflow and downstream applications, thus we expect them to be decent critics of generation quality.

In the study, we present four sets of textured objects: an ambulance, a dumpster, a compass, and a backpack, via online questionnaires. Each set contains four textures generated by CLIPMesh, Latent-Paint, Text2Tex~\cite{chen2023text2tex}, and our method. The participants are asked to rank the results in terms of generation quality, taking into account correctness, fidelity, level of detail and overall aesthetics.

\begin{figure}[h]
    \centering
    \includegraphics[width=0.9\linewidth]{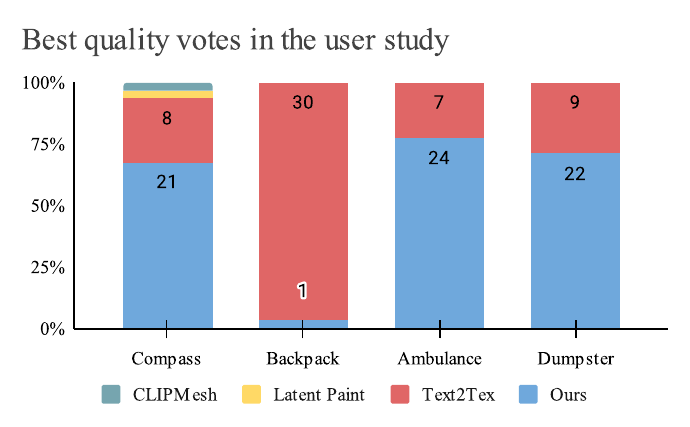}
    \caption{\textbf{Vote distribution for best quality.} Ours was selected as the best for most of the objects.}
    \label{fig:user1}

    \centering
    \includegraphics[width=0.9\linewidth]{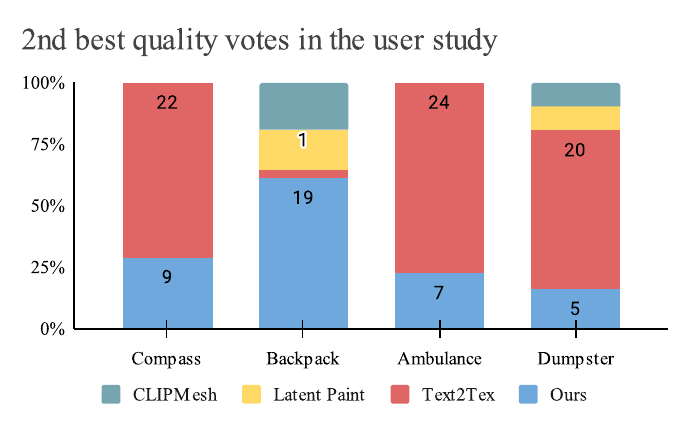}
    \caption{\textbf{Vote distribution for second best quality.} SDS-based methods overall received more votes than the CLIP-based ones.}
    \label{fig:user2}
\end{figure}

The results are shown in Figure~\ref{fig:user1} and ~\ref{fig:user2}. Most participants selected their top 2 choices from either SOTA Text2Tex~\cite{chen2023text2tex} or ours. For the ambulance and dumpster objects, the votes are unanimously toward our generated textures. For the compass, about two-thirds voted for ours. The votes for the backpack are diverse for the second-best quality object. 

Overall, the study indicates a high acceptance of the textures generated by our model.

\subsection{Convergence Speed}

\begin{figure}
    \centering
    \includegraphics[width=0.9\linewidth]{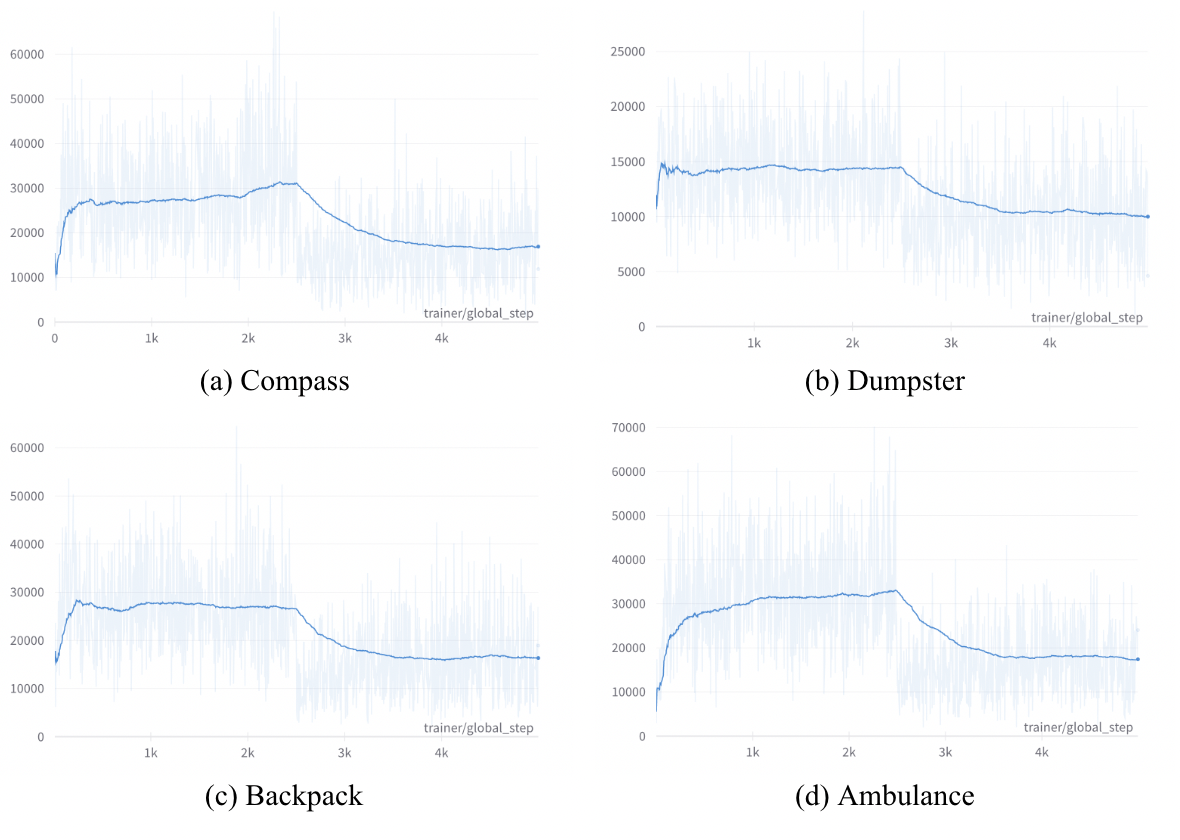}
    \caption{\textbf{The SDS loss throughout the iterations.} It takes Euclidreamer around 4300 steps, and even 2500 steps gives decent result already.}
    \label{fig:loss}
\end{figure}

Figure~\ref{fig:loss} shows the SDS loss throughout the training steps. In general, it takes EucliDreamer around 4300 steps to converge the loss. Note that for all 4 objects being tested, most progress is already made before 2500 steps. In contrast, under identical experimental conditions and using the same set of hyperparameters, DreamFusion~\cite{poole2022dreamfusion} requires over 10,000 steps to converge. Thus, the runtime of EucliDreamer is about half of that of DreamFusion. This comparison highlights the superior convergence speed and computational efficiency of our method.

\subsection{Importance of Depth Conditioning}

\begin{figure}
    \centering
    \includegraphics[width=\linewidth]{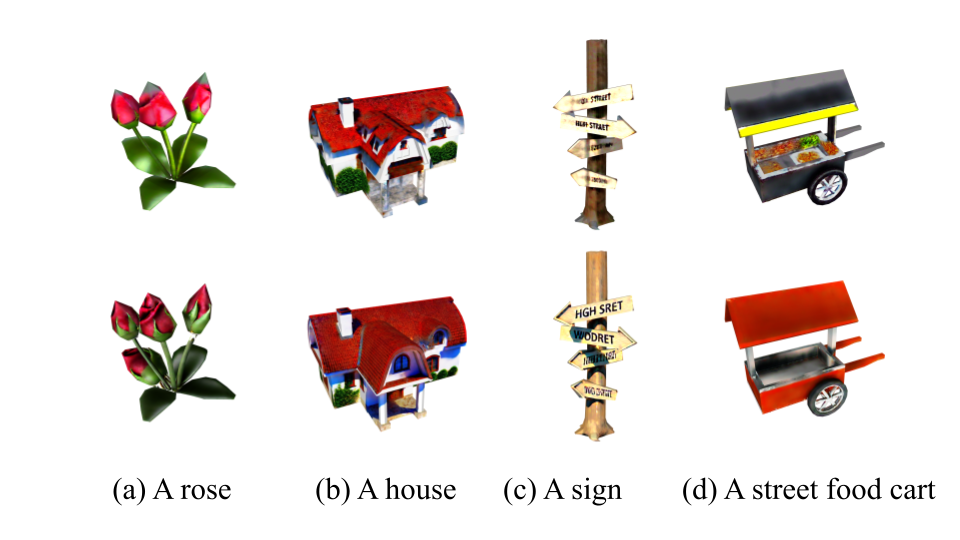}
    \caption{\textbf{The effect of depth conditioning.} The top row does not include depth conditioning, and the bottom row does. With depth conditioning: (a) The rose buds shows more detailed textures. (b) The house has its windows and doors correctly placed. (c) The sign has more detailed texts. (d) The cart shows semantically correct textures, i.e. no food on the container wall.}
    \label{fig:problem_fix}
\end{figure}

Figure \ref{fig:problem_fix} compares the texturing results on 4 models with and without depth-conditioning. We observe that depth-conditioning is especially helpful for ensuring correctness. For instance, without depth information, windows can show up on the roof of the house, and food texture can incorrectly appear on the wall of the food cart container. In addition, depth-conditioning seems to improve the detail quality and texture sharpness. This might due to that it removes a lot of ambiguities and prevents many modes from mixing together.


\subsection{Various Art Styles}

Our second set of experiments focuses on the diversity of the generated textures. By providing input text prompts of the model content and styles, one can use EucliDreamer to generate 3D textures for a given 3D mesh in certain styles. For example, a prompt of "a building, damaged, dirty, 3D rendering, high-quality, realistic" will give the building model a rusty and pale feeling, as shown in Figure~\ref{fig:styles} in Supplementary Materials. Through simply alternating the text prompts, EucliDreamer can provide diverse combinations of color tones, brush strokes, lighting and shadows in the textures, which highlights its flexibility and "creativity".

\section{Discussion}
\label{sec:dicussions}

Through the experiments, we have proved that our method using depth-conditioned Stable Diffusion can generate higher-quality textures in a shorter time. This is as expected, for the following reasons:

In terms of quality, depth information further restricts the probability density on top of the text prompts, eliminating unnatural or physically impossible cases. In some cases, without the depth guidance, the model may generate a texture for a human character with multiple faces, or a tree with leaf textures on where is supposed to be the stem.

In terms of generation time, the time it takes for our method to convergence is shorter for two reasons. First, depth conditioning serves as a constraint that eliminates many possibilities, preventing the "back and forth" between different modes. Second, we define the implicit function representing the texture such that it sticks to the surface of the object. Both led to a smaller probability space that takes less time to search.

More complex methods exist to address the mode collapse problem and enhance texture quality, e.g. using VSD ~\cite{wang2023prolificdreamer} instead of SDS. However, as shown in our study in Supplementary Materials, for generating textures, using depth conditioning achieves comparable quality while being much simpler.



\subsection{Limitations and Future Works}

Our method, like many other SDS-based methods, requires that all surfaces of an object to be visible from some angles. Therefore, it may encounter issues with extremely complex structures with holes or obstructions. This might be mitigated by splitting the model into parts. 

Also, as our method relies on text-to-image diffusion models that were trained to produce realistic natural images, which may produce "baked-in" effects into the texture, such as shadows and highlights. This may be undesirable for use cases that requires re-lightable 3D assets. As a future work, a possible solution is to fine-tune a Stable Diffusion variant that produces images with flat lighting style.

\subsection{Conclusion}
We present a simple and effective solution for automatically generating textures for 3D meshes. High-quality texturing is one of the labor-intensive stages in 3D modeling workflow, which can take an experienced artist many days to complete for a single object. We hope our method can reduce the cost for preparing 3D assets in applications like gaming, animation, virtual reality \cite{wu2014real, joo2018total, congruiNVS2023} or even synthetic training data \cite{song2023synthetic, qiu2021synface, congruiPSCZ2022} for computer vision models \cite{congruiSAMRoad2024, congruiRQEN2018, congruiImp2017, congruiDD2019}, and inspire further research in this direction.

\nocite{KaolinLibrary,ravi2020accelerating,hertz2023delta,sella2023voxe,armandpour2023reimagine,an2023panohead}

\nocite{congruiCR2023, congruiSliding2023, congruiUnlock2023, congruiPImp2021, congruiPOpt12023, congruiPOpt22022, congruiPStop2023, congruiSAMRoad2024}
{
    \small
    \bibliographystyle{ieeenat_fullname}
    \bibliography{main}
}

\clearpage
\maketitlesupplementary

\setcounter{page}{1}

\section{Implementation details}
\label{sec:suppl1}

To better demonstrate our approach and reproduce our results, we will publish our code on GitHub upon acceptance.


\subsection{Parameter selection}
\label{sec:parameters}
For the main experiment, we selected the following parameters that performed the best in ~\ref{sec:ablation}. The Adam optimizer has a learning rate of 0.01.

\begin{table}[!ht]
    \centering
    \begin{tabular}{|l|l|}
    \hline
        \textbf{Parameter} & \textbf{Value} \\ \hline
        Batch size & 8 \\ \hline
        Camera distance & [1.0, 1.5] \\ \hline
        Elevation range & [10, 80] \\ \hline
        Sampling step & [0.02, 0.98] \\ \hline
        Optimizer & Adam \\ \hline
        Training steps & 5000 \\ \hline
        Guidance scale & 100 \\ \hline
    \end{tabular}
\end{table}

The prompt keywords below are added in addition to the original phrase of the object (e.g.  “a compass”).

\begin{verbatim}
prompt:
  "animal crossing style, a house,
  cute, Cartoon, 3D rendering, red
  tile roof, cobblestone exterior"
negative_prompt:
  "shadow, greenshadow, blue shadow,
  purple shadow, yellow shadow"
\end{verbatim}

The texture resolution is set to 512*512.

\subsection{Infrastructure}

ThreeStudio

(https://github.com/threestudio-project/threestudio)

is a unified framework for 3D modeling and texturing from various input formats including text prompts, images, and 3D meshes. It provides a Gradio backend with easy-to-use frontend with input configurations and output demonstration.

We forked the framework and added our approach with depth conditioning. A depth mask is added to encode the depth information. 

\begin{verbatim}

with torch.no_grad():
    depth_mask =
      self.pipe.prepare_depth_map(
        torch.zeros((batch_size, 3,
          self.cfg.resolution,
          self.cfg.resolution)),
        depth_map,
        batch_size=batch_size,
        do_classifier_free_guidance=True,
        dtype=rgb_BCHW.dtype,
        device=rgb_BCHW.device,
    )

\end{verbatim}

\subsection{Hardware}

We used a Nvidia RTX 4090 GPU with 24GB memory. This allows us to run experiments with a batch size up to 8 and dimension 512*512. In theory, a GPU with more memory will allow us to generate textures with higher definition and in better quality.

\section{More results from Ablation Studies}
\label{sec:suppl}

Due to the page limit, we show experimental results of elevation range (section~\ref{sec:elevation}), batch size (section~\ref{sec:parameters}), guidance scale (section~\ref{sec:guidance}), negative prompts (section~\ref{sec:negative}), and data augmentation (section~\ref{sec:augmentation}) here. The results support our conclusions in the Ablation Studies section~\ref{sec:ablation}.

For model understanding and parameter selection, we conduct a series of experiments for EucliDreamer with different settings.


\begin{figure*}
    \centering
    \includegraphics[width=0.8\textwidth]{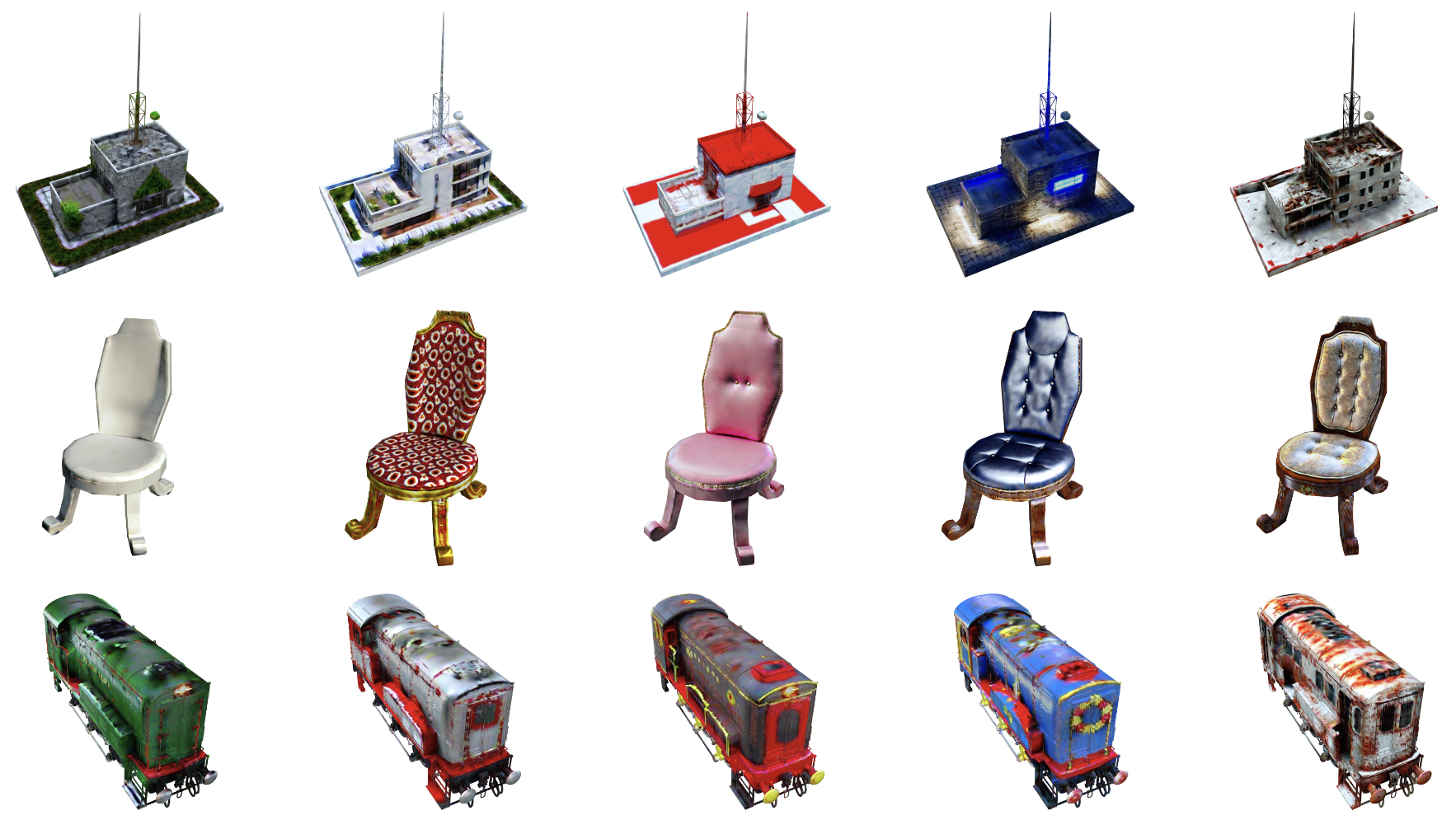}
    \caption{Objects textured by EucliDreamer in various styles, by different input text prompts.}
    \label{fig:styles}
\end{figure*}



\subsection{SDS vs. VSD}

\begin{figure*}
    \centering
    \includegraphics[width=0.7\textwidth]{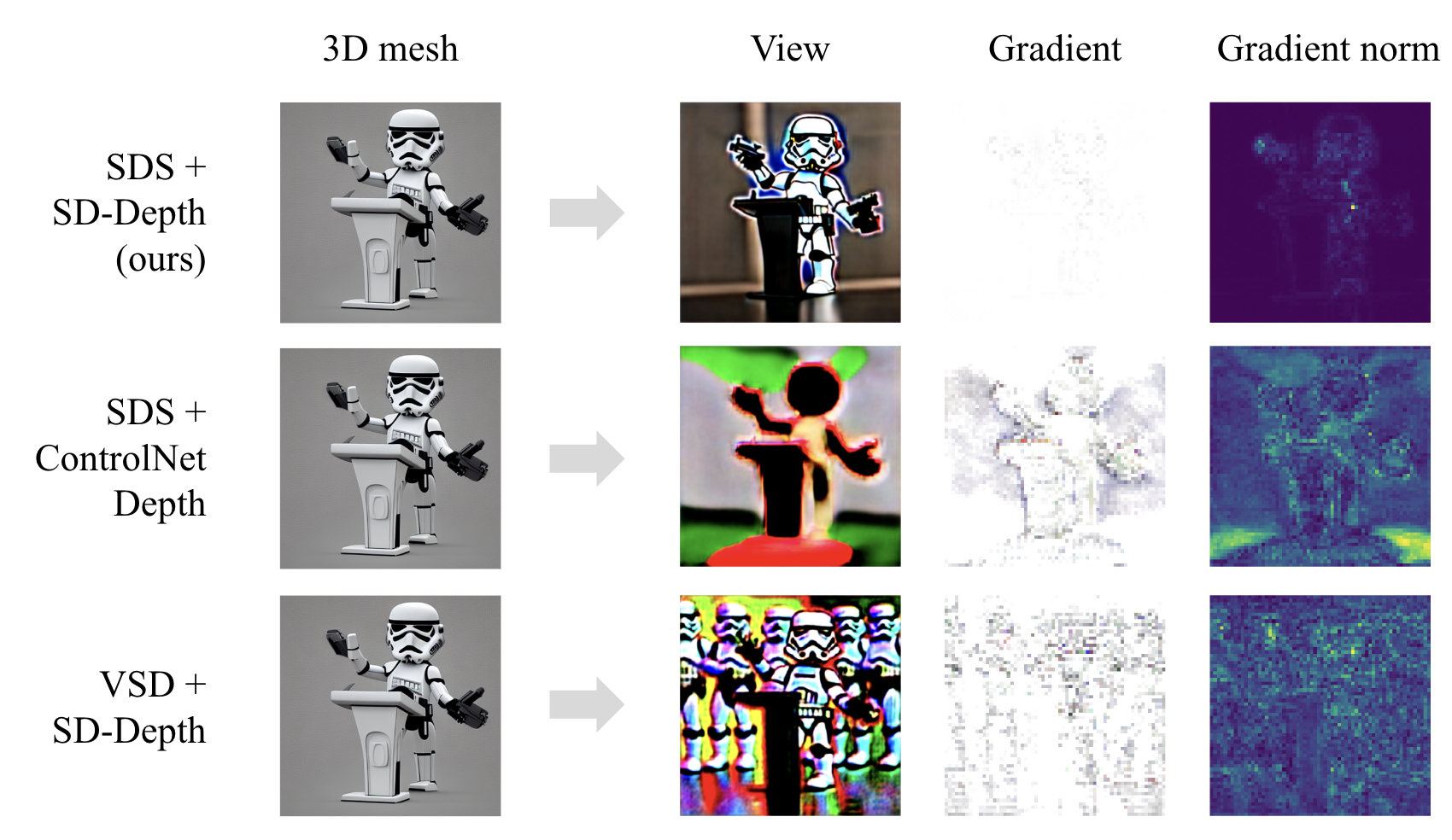}
    \caption{\textbf{Comparisons among multiple depth techniques.} The SDS and ControlNet depth combination generates a view with missing details of the character. The VSD~\cite{wang2023prolificdreamer} and Stable Diffusion depth combination leads to multiple characters and noisy gradient results. Overall, Our method with SDS and Stable Diffusion depth generates the best gradient for the given mesh.}
    \label{fig:controlnet}
\end{figure*}

SDS~\cite{poole2022dreamfusion} is notoriously known for its lack of diversity and overly smoothed texture. This is because SDS is mean-seeking. Intuitively, a single prompt has multiple plausible generation results. SDS tends to create an average of them (mean seeking). Variational Score Distillation (VSD) is proposed in ProlificDreamer~\cite{wang2023prolificdreamer} to mitigate this problem by introducing a variational score that encourages diversity. However, in depth-conditioned texturing, there are fewer plausible generation results given a single prompt and depth. This leads to the diversity and details in our approach (mode seeking), even though VSD is not used. Through experiments shown in Figure~\ref{fig:controlnet}, our approach produces better results without VSD. Thus VSD in ProlificDreamer~\cite{wang2023prolificdreamer} is not required when Stable Diffusion depth is used.

\subsection{StableDiffusion depth vs. ControlNet depth}

The most straightforward way of depth-conditioned SDS is to use ControlNet~\cite{zhang2023adding}. However, empirically we found that the SDS gradient from ControlNet~\cite{zhang2023adding} is much noisier than SD-depth. SD-depth concatenates depth with latents. The SDS gradients flow to depth and latents separately, which is less noisy. As shown in Figure~\ref{fig:controlnet}, we demonstrate better generation results with Stable Diffusion depth than ControlNet depth. 

\subsection{Elevation range}
\label{sec:elevation}
Elevation range defines camera positions and angles at the view generation step. When it is set between 0-90 degrees, cameras face down and have an overhead view of the object. We find that fixed-angle cameras may have limitations and may miss certain angles, resulting in blurry color chunks at the surface of the object. One way to leverage the parameter is to randomize camera positions so all angles are covered to a fair level.

\subsection{Sampling min and max timesteps}

The sampling min and max timesteps refer to the percentage range (min and max value) of the random timesteps to add noise and denoise during SDS process. The minimum timestep sets the minimum noise scale. It affects the level of detail of the generated texture. The maximum timestep affects how drastic the texture change on each iteration or how fast it converges. While their values may affect quality and convergence time, no apparent differences were observed between different value pairs of sampling steps. In general, 0 and 1 should be avoided for the parameters.

\subsection{Learning Rate}
\label{sec:learningrate}
In the context of 3D texture generation, a large learning rate usually leads to faster convergence. A small learning rate creates fine-grained details. We found learning rate of 0.01 with the Adam optimizer is good for 3D texture generation. 

\subsection{Batch Size}
\label{sec:batchsize}
We set batch sizes to 1, 2, 4, and 8 respectively. Results show that a larger batch size leads to more visual details and reduces excessive light reflections and shadows. More importantly, large batch size increases view consistency. This is because gradients from multiple views are averaged together and the texture is updated once. A batch size of 8 is enough from our observation. A batch size of more than 8 does not lead to a significant reduction in the number of iteration steps. 

\subsection{With vs. without gradient clipping}

Intuitively gradient clipping can avoid some abnormal updates and mitigate the Janus problem, shown in Debiased Score Distillation Sampling (D-SDS)~\cite{hong2023debiasing}. In the context of texture generation, we do not observe Janus problem due to the fact that the mesh is fixed. In our experiments, there's no noticeable benefit of gradient clipping. 

\subsection{Guidance scale}
\label{sec:guidance}
The guidance scale specifies how close the texture should be generated based on input text prompts. We find that the guidance scale affects diversity and saturation. Even with the depth conditional diffusion model, a high guidance scale like 100 is still required. 

\subsection{Negative prompts}
\label{sec:negative}
Negative prompts in Stable Diffusion can fix some artifacts caused by SDS. Adding negative prompts like "shadow, green shadow, blue shadow, purple shadow, yellow shadow" will help with excessive shadows.

\subsection{Data augmentation}
\label{sec:augmentation}

\begin{figure}
    \centering
    \includegraphics[width=0.45\textwidth]{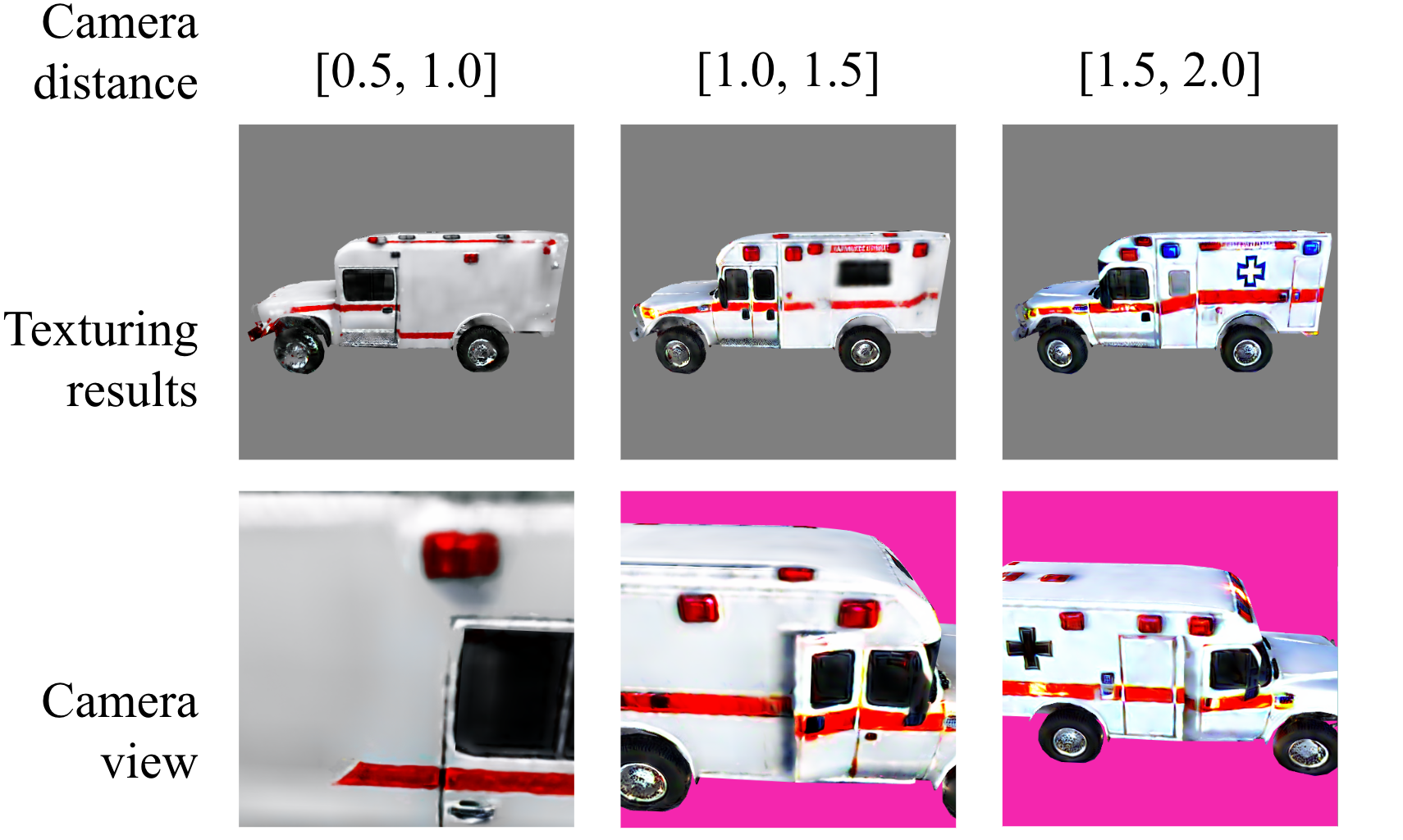}
    \caption{\textbf{Comparisons among camera distance values.} The one with [1.0, 1.5] has the highest quality.}
    \label{fig:camera}
\end{figure}

We found that adjusting the camera distance can improve both resolution and generation quality. We experimented with three different ranges of camera distance, shown in Figure~\ref{fig:camera}. The distance range of [1.5, 2.0] produces the optimal outcomes regarding the quality of textures and colors. This can be attributed to its ability to simulate viewpoints most representative of conventional photography, similar to the usual perspectives from which an ambulance is observed.

\subsection{Image-to-texture using Dreambooth3D}
\begin{figure}
    \centering
    \includegraphics[width=0.45\textwidth]{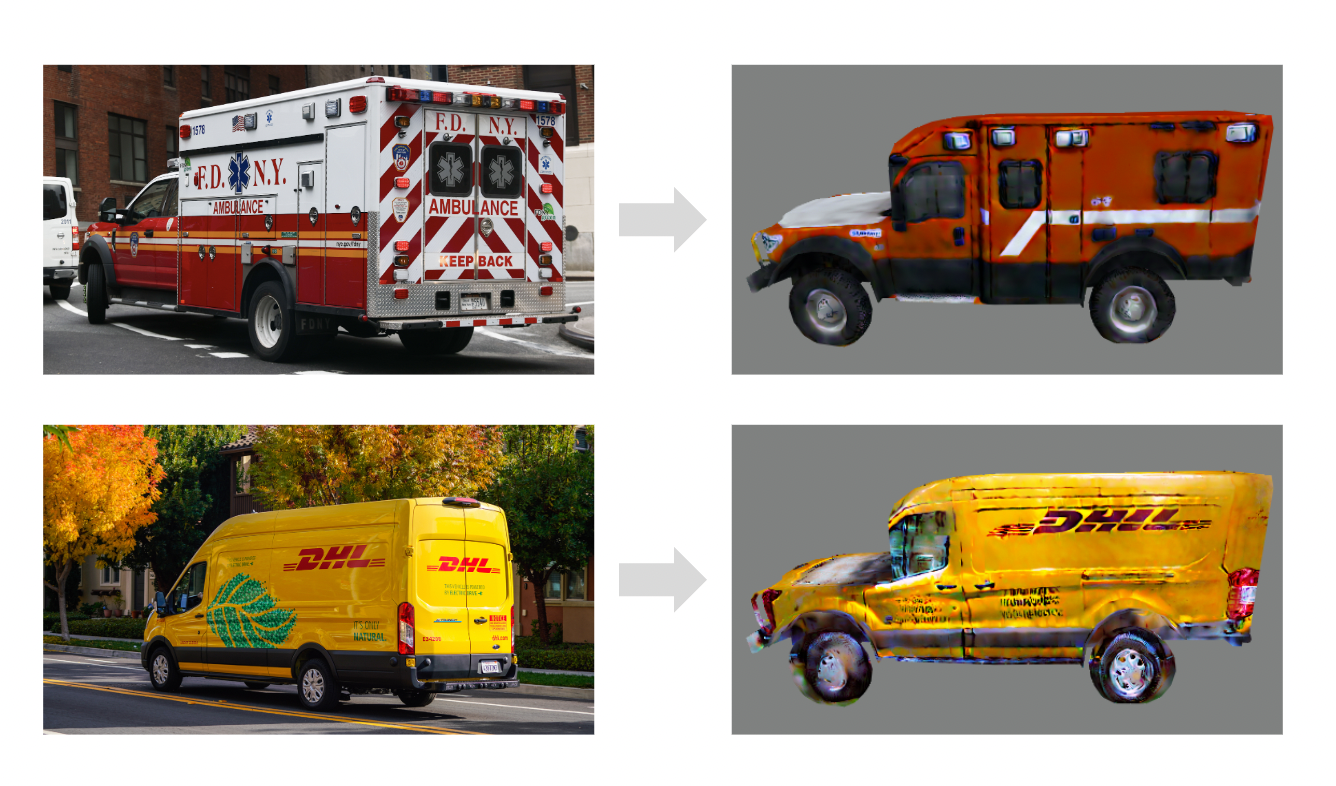}
    \caption{\textbf{Image-to-texture generation using Dreambooth3D~\cite{wang2023prolificdreamer}.} The generated textures have the same color tones as the input images.}
    \label{fig:dreambooth}
\end{figure}

Dreambooth3D~\cite{raj2023dreambooth3d} was initially proposed as an extension of DreamFusion~\cite{poole2022dreamfusion} to enable image-to-3D shape generation. We partially finetuned Stable Diffusion depth with the user-provided image(s), generated textures with finetuned stable-diffusion-depth from step 1, fine-tuned Stable Diffusion depth again with outputs from step 2, and then generated final results with finetuned stable-diffusion-depth from step 3. Shown in Figure~\ref{fig:dreambooth}, combined with Dreambooth3D~\cite{raj2023dreambooth3d}, our approach generates satisfactory textures given a 3D mesh and a single image as inputs. 

\begin{figure}[b!]
    \centering
    \includegraphics[width=0.45\textwidth]{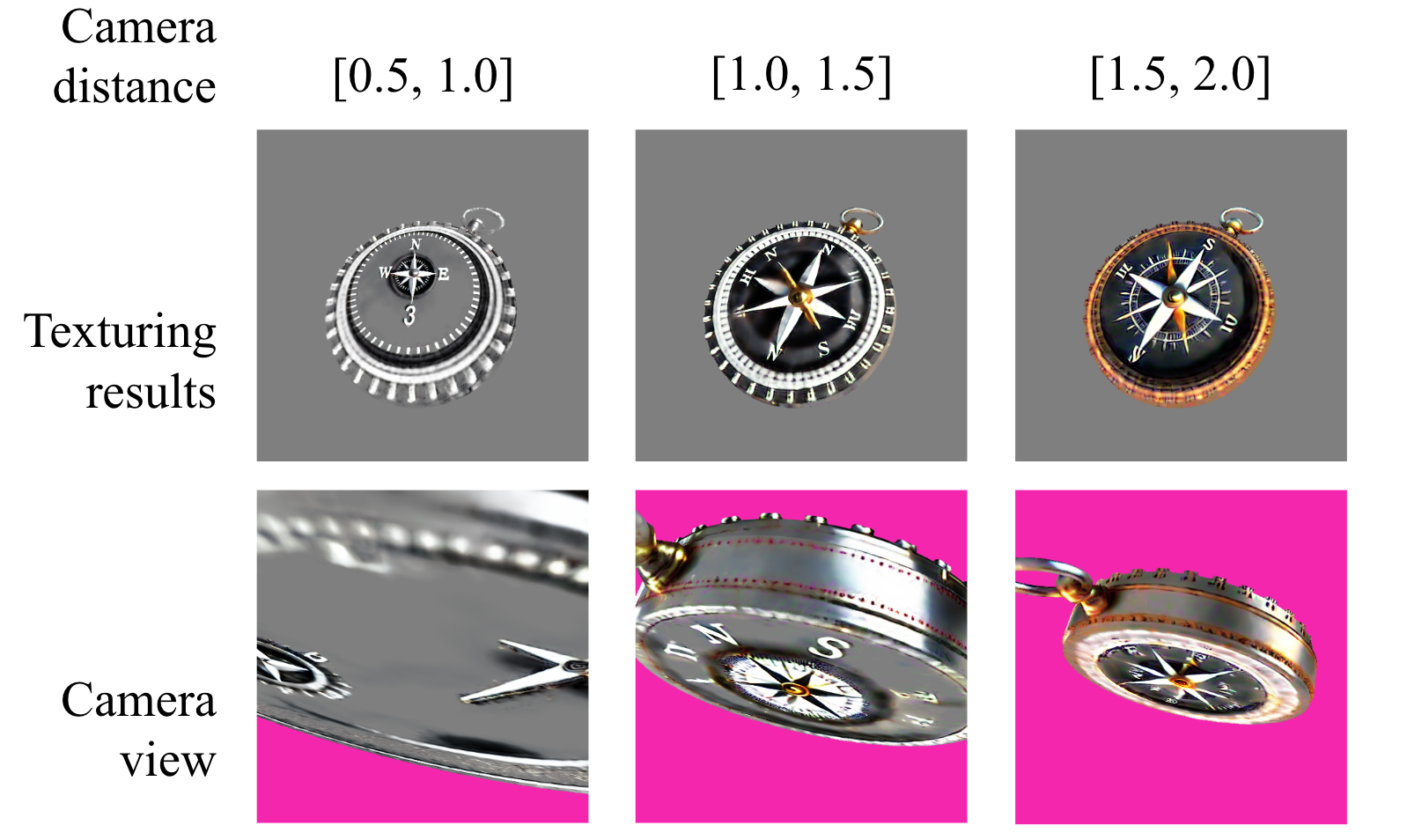}
    \caption{\textbf{Additional comparisons among camera distance values.} The one with [1.0, 1.5] has the highest quality for compass.}
    \label{fig:camera2}
\end{figure}

\begin{figure*}
    \centering
    \includegraphics[width=0.8\textwidth]{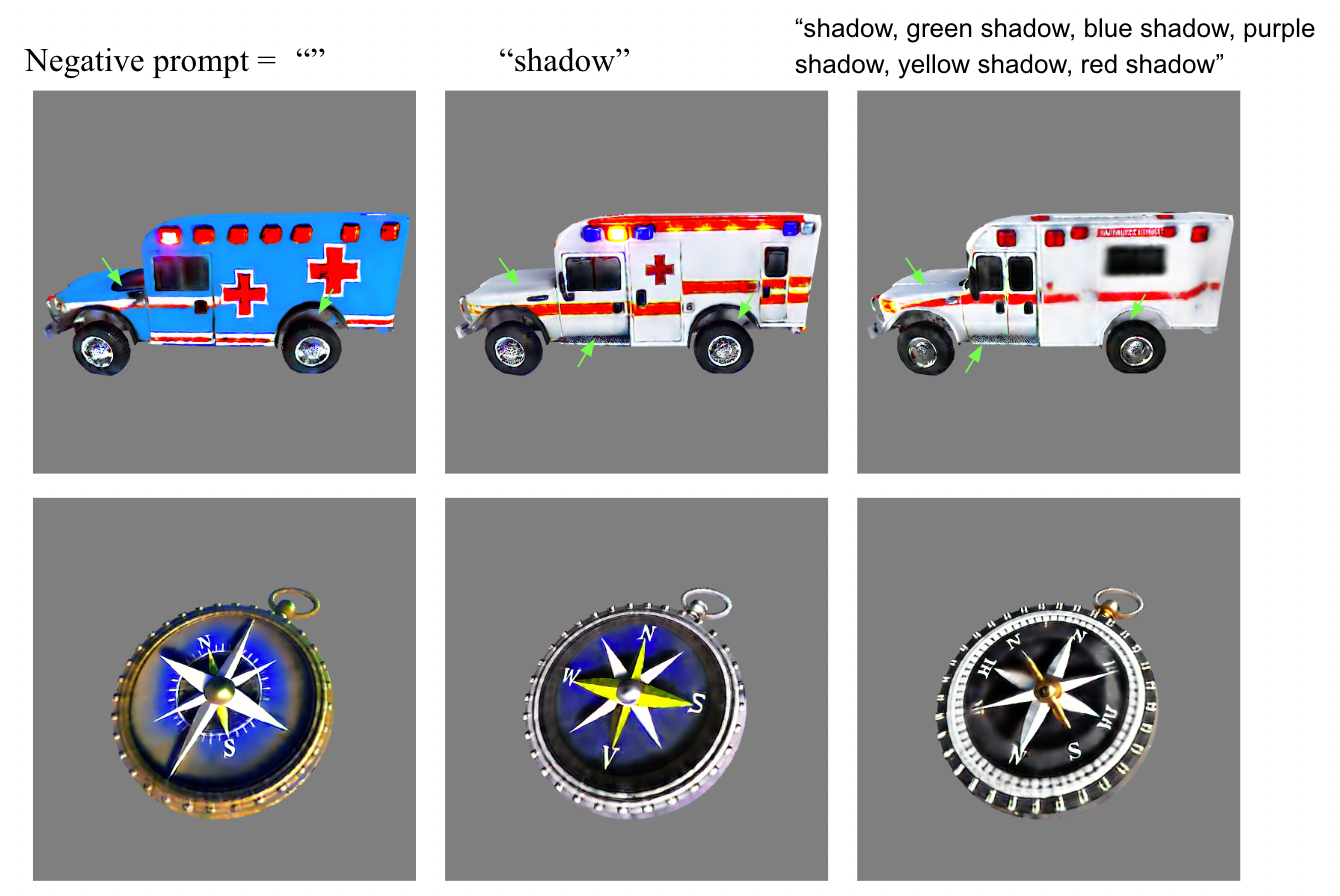}
    \caption{\textbf{Study on negative prompt.} Empty negative prompts yield weird coloring and significant baked-in shadow (see the hood). Negative prompts can be used for shadow suppression. Specifying a wide coverage removes more shadow.}
    \label{fig:neg_prompt}
\end{figure*}

\begin{figure*} [t!]
    \centering
    \includegraphics[width=0.8\textwidth]{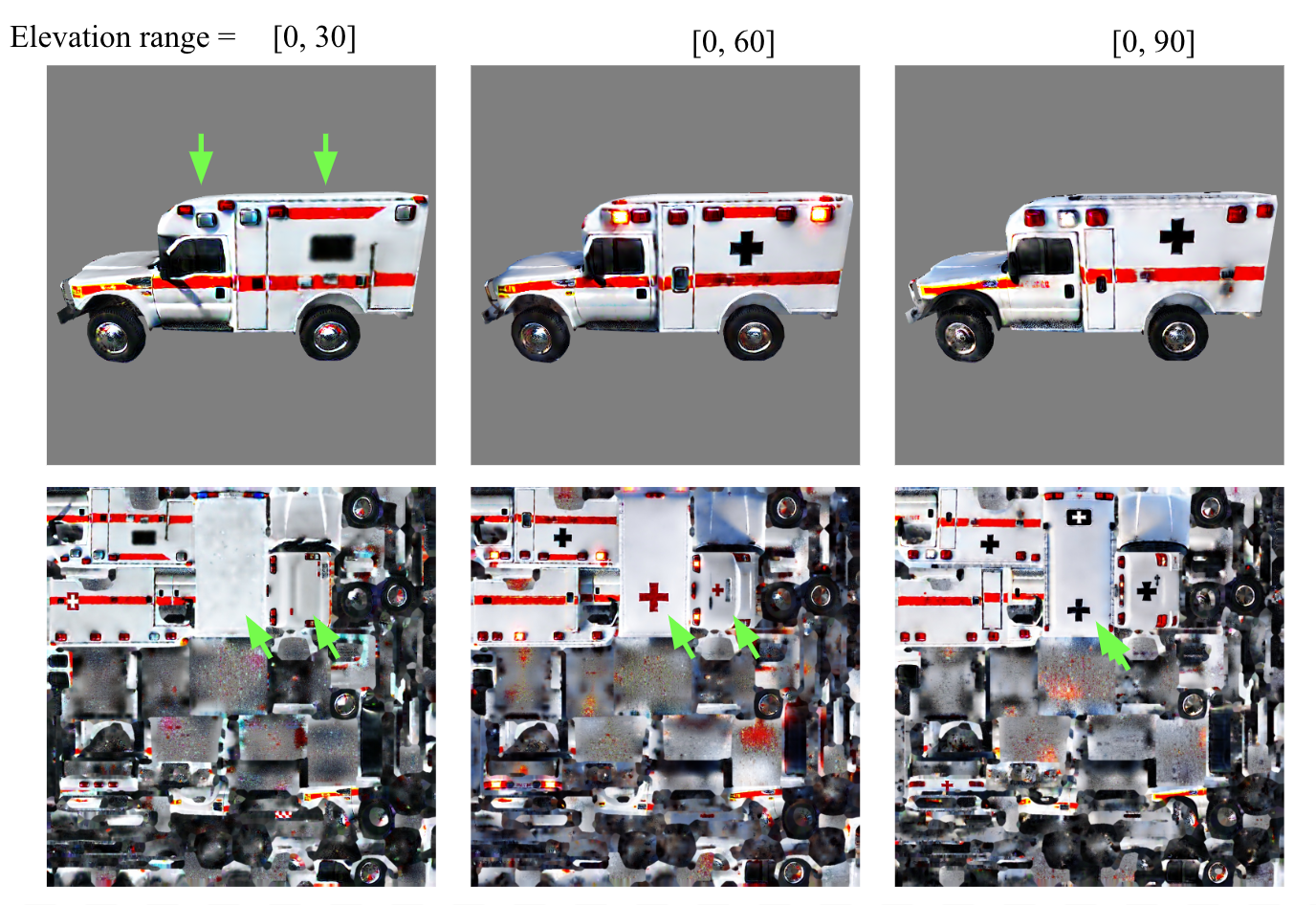}
    \caption{\textbf{Study on elevation range.} A wider camera elevation range gives better texture coverage. For example, as the elevation upper bound increases, more texture details show up on the top surfaces.
}
    \label{fig:ele_range}
\end{figure*}

\begin{figure*} [h!]
    \centering
    \includegraphics[width=0.8\textwidth]{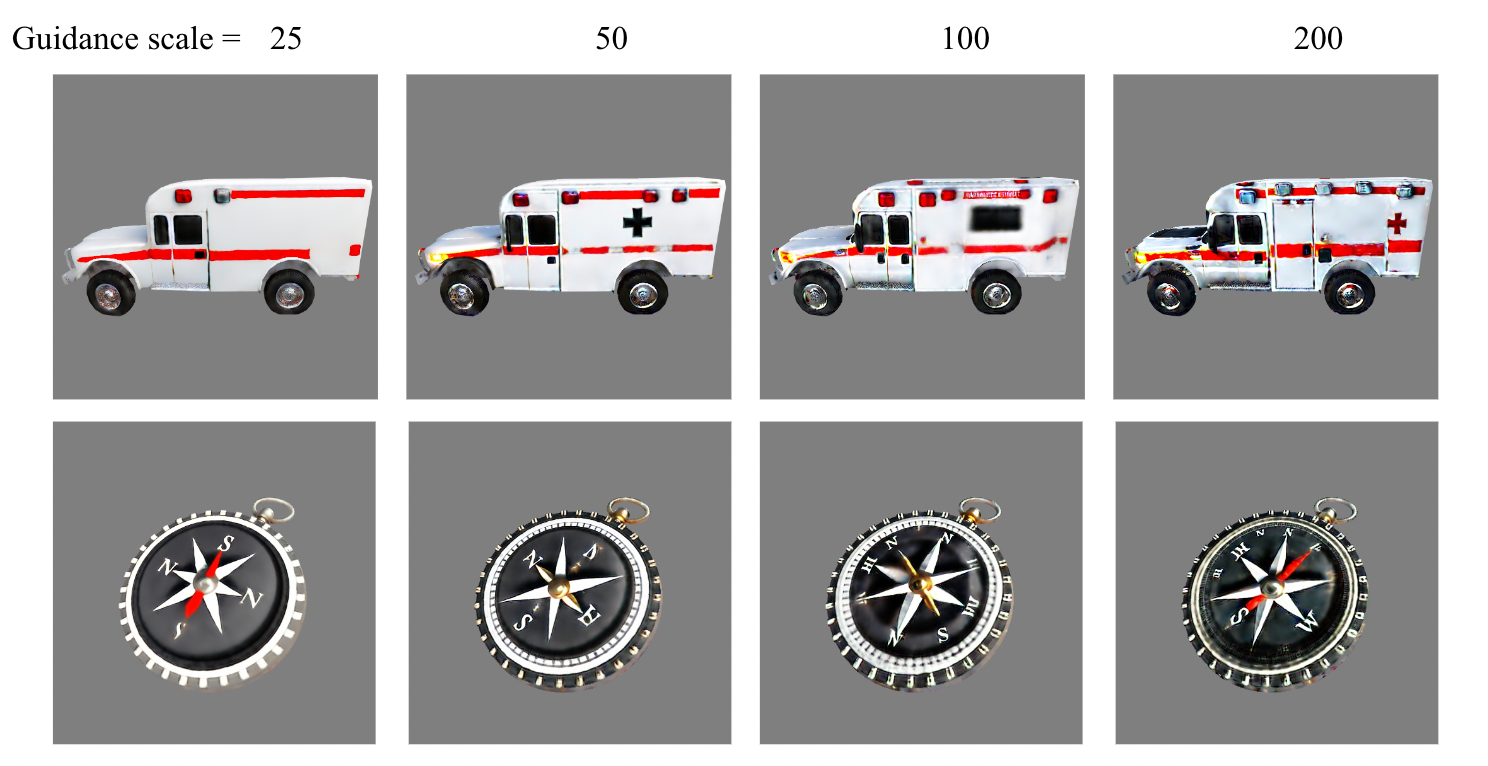}
    \caption{\textbf{Study on guidance scale.} It seems that as the guidance scale increases, more details are added, and “photorealism” seems to increase. However, it seems the noise level also increases.
}
    \label{fig:guidance}
\end{figure*}

\begin{figure*}
    \centering
    \includegraphics[width=0.8\textwidth]{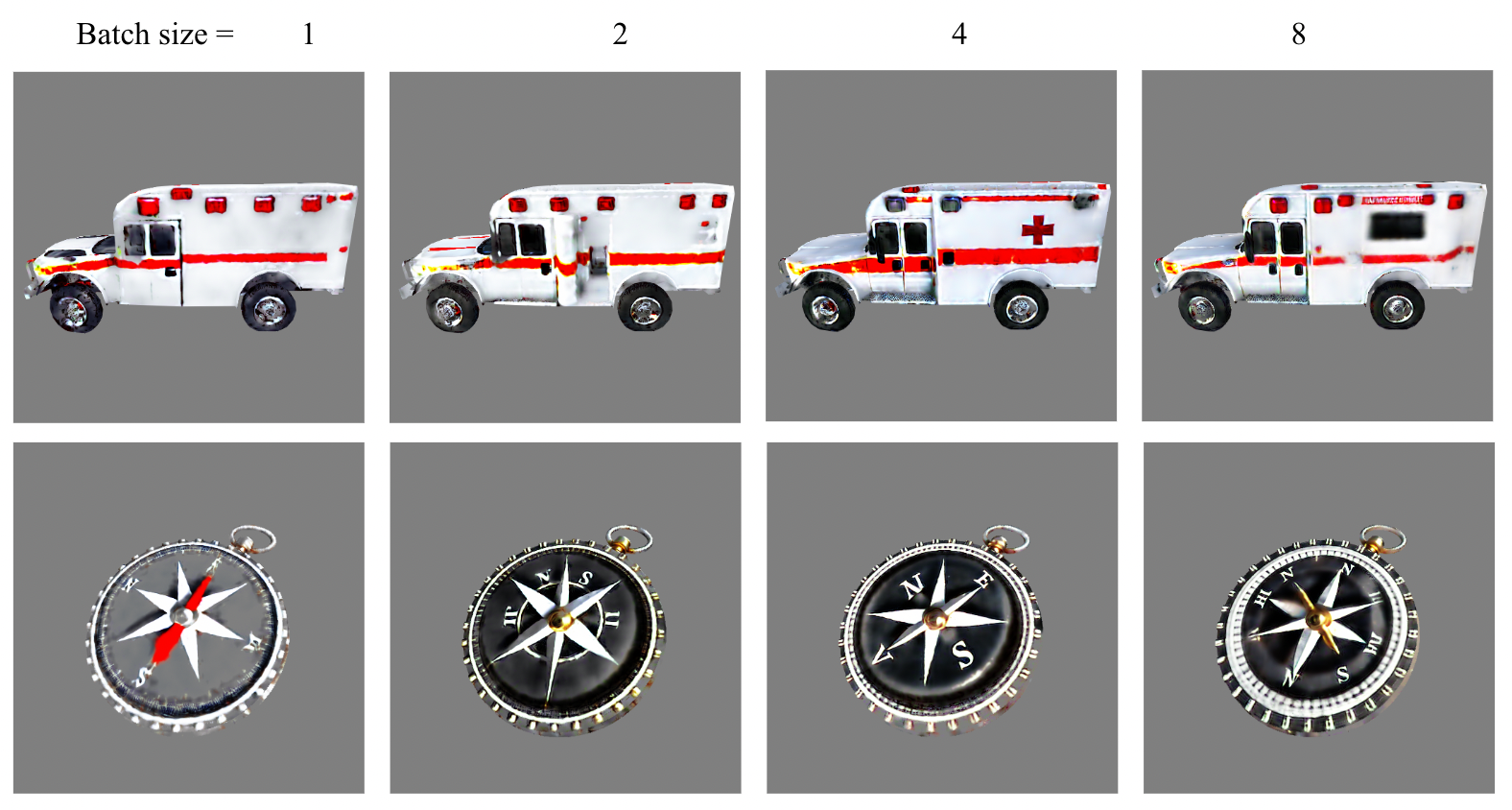}
    \caption{\textbf{Study on batch size.} Texture quality improves as batch size increases.}
    \label{fig:batch_size}
\end{figure*}

\end{document}